\documentclass[10pt,twocolumn,letterpaper]{article}

\usepackage{wacv}
\usepackage{times}
\usepackage{epsfig}
\usepackage{graphicx}
\usepackage{amsmath}
\usepackage{amssymb}
\usepackage{csquotes}
\usepackage{graphicx}
\usepackage{rotating}
\usepackage{multirow}
\usepackage[normalem]{ulem}
\usepackage{algorithm}
\usepackage{algorithmicx}
\usepackage{algpseudocode}
\usepackage{booktabs}
\usepackage{array}
\usepackage{rotating}
\usepackage{arydshln}
\usepackage{ragged2e}
\usepackage{multirow}
\usepackage[normalem]{ulem}
\usepackage{tabularx}

\def\wacvPaperID{975} 

\wacvfinalcopy 

\ifwacvfinal
\usepackage[breaklinks=true,bookmarks=false]{hyperref}
\else
\usepackage[pagebackref=true,breaklinks=true,colorlinks,bookmarks=false]{hyperref}
\fi

\begin{document}

\title{Inferring the Class Conditional Response Map \\for Weakly Supervised Semantic Segmentation}

\author{Weixuan Sun,   Jing Zhang,   Nick Barnes\\
Australian National University\\
{\tt\small \{weixuan.sun, jing.zhang, nick.barnes\}@anu.edu.au}
}

\maketitle

\begin{abstract}
Image-level weakly supervised semantic segmentation (WSSS) relies on class activation maps (CAMs) for pseudo labels generation. As CAMs
only highlight the most discriminative regions of objects, the generated pseudo labels are usually unsatisfactory to serve directly as supervision. To solve this, most existing approaches follow a multi-training pipeline to refine CAMs for better pseudo-labels, which includes: 1) re-training the classification model to generate CAMs; 2) post-processing CAMs to obtain pseudo labels; and 3) training a semantic segmentation model with the obtained pseudo labels. However, this multi-training pipeline requires complicated adjustment and additional time. To address this, we propose a class-conditional inference strategy and an activation aware mask refinement loss function to generate better pseudo labels without re-training the classifier. The class conditional inference-time approach is presented to separately and iteratively
reveal the classification network's hidden object activation to generate more complete response maps. Further, our activation aware mask refinement loss function introduces a novel way to exploit saliency maps during segmentation training and refine the foreground object masks without suppressing background objects.
Our method achieves superior WSSS results without requiring re-training of the classifier. \url{https://github.com/weixuansun/InferCam}

\end{abstract}
\vspace{-5mm}

\section{Introduction}
Recent work on 2D image semantic segmentation has achieved great progress via
deep fully convolutional neural networks (FCN) \cite{long2015fully}. 
The success of these models \cite{zhao2017pyramid,chen2017deeplab,chen2014semantic,zhao2018psanet,chen2017rethinking} comes from large training datasets with pixel-wise labels, which are laborious and expensive to obtain. 
To relieve the labeling burden, multiple types of weak labels have been studied, including image-level \cite{huang2018weakly,wei2018revisiting,ahn2018learning,fan2020cian}, points \cite{bearman2016s},
scribbles \cite{vernaza2017learning,lin2016scribblesup,tang2018regularized}, and bounding boxes \cite{dai2015boxsup,papandreou2015weakly,khoreva2017simple,li2018weakly,song2019box}.
In this paper, we focus on weakly-supervised semantic segmentation with image-level labels, the lowest annotation-cost alternative.

\begin{figure}[t!]
   \begin{center}
   {\includegraphics[width=1.0\linewidth]{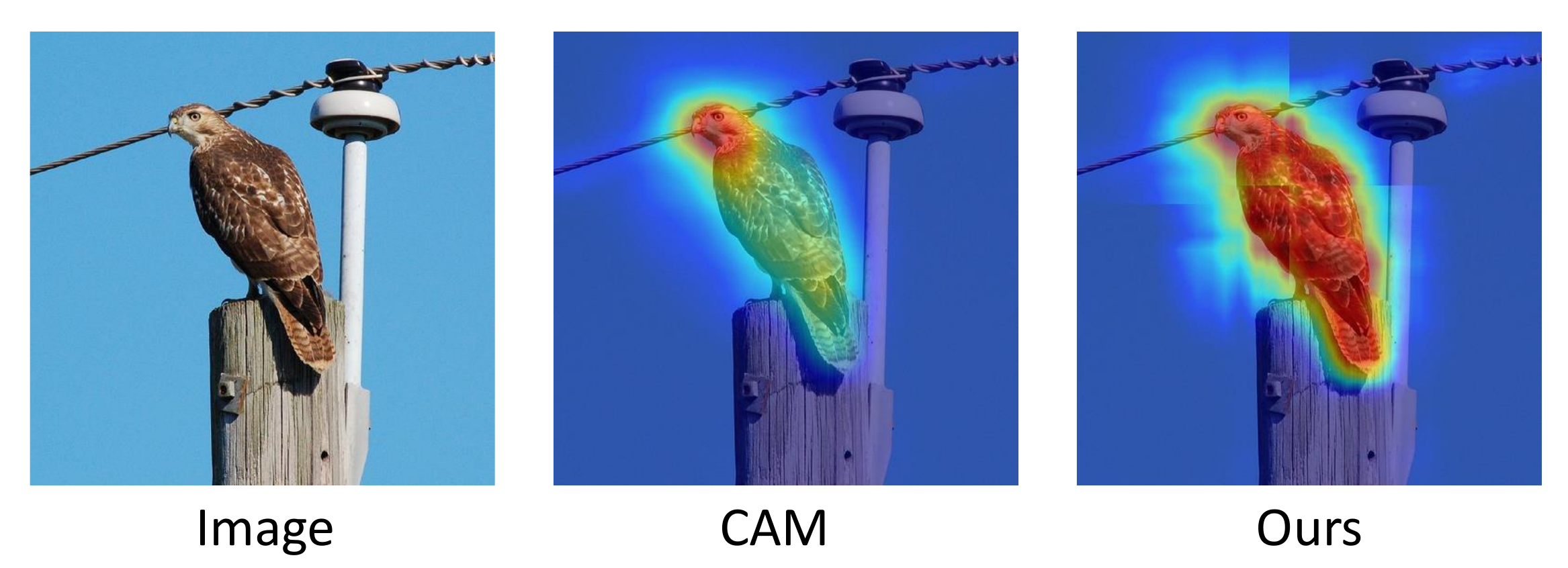}}
   \end{center}
\caption{Our  object response maps compared with baseline CAMs. 
The baseline CAMs only highlight the most discriminative regions. 
Our proposed
technique
leads to response maps that 
integrally 
cover larger object regions without re-training.}
\vspace{-5mm}
   \label{fig: introduction}
\end{figure}


The typical way to learn from image-level labels usually involves progressive steps: 1) an initial response map (the class activation map
(CAM) 
\cite{zhou2016learning}) 
is obtained
to roughly locate the objects; 2) 
pseudo labels are generated based on the initial response map with post-processing techniques,
\eg, denseCRF \cite{krahenbuhl2011efficient}, random walk \cite{lovasz1993random} or an additional network\cite{ahn2019weakly, ahn2018learning, wang2018weakly}; 3) a semantic segmentation network is trained
with the pseudo labels as supervision. 
The quality of the initial response map plays an important role
for image-level WSSS,
as good
response maps can
fill the inherent gap between image-level labels and pixel-wise labels.
However,
as the CAM highlights discriminative regions of each category, the partial activation
leads to unsatisfactory semantic segmentation.
To refine the pseudo labels from CAMs, 
most recent state-of-the-art methods require additional training steps.
\ie re-train the classification model to encourage the CAMs to cover more object areas \cite{chang2020mixup, chang2020mixup, wei2018revisiting, sun2020mining, zhang2020splitting, wang2020self}, or train additional networks \cite{ahn2018learning, ahn2019weakly, wang2018weakly} to guide the CAMs to generate pseudo labels.


We observe that
the partial activation of the CAM is because only the discriminative region is usually needed for effective object classification.
So the network is prone to focus on the discriminative areas.
However, 
we argue that this does not indicate that the classifier learns nothing about other less-discriminative patterns. 
We experimentally validate that,
assuming there is
sufficient data for each category, the trained classifier can generate activation on most areas of objects but unevenly distributed. 
We describe this partial activation issue as an \enquote{unequal distribution of the activation}, and find that
conventional inference fails to leverage the full power of the baseline classifier. We demonstrate
that
a basic classification network, pre-trained on the target dataset without
modification is sufficient
to generate uniform activation and cover most object areas with an effective inference strategy.

We propose an inference-time 
image augmentation method
to reveal hidden activation of
objects and generate better object response maps for WSSS.
To prevent the risk of diverging from the well-trained classifier, we do not
re-train the baseline classifier. Instead, our method adopts only a novel inference mechanism to deal with the unequal distribution issue of the CAM.
Specifically, we first introduce a \enquote{split \& unite} based image augmentation method to encourage the network to pay attention to different parts of objects and generate equal activation on each part.
To further push the activation to other less discriminative areas, we present a \enquote{hide and re-inference} method, which iteratively mines activated regions of objects and aligns them to an equally-distributed response map. 
Finally, we integrate these two modules into a simple framework that can be used in the inference stage of existing classifiers.
As shown in Fig.~\ref{fig: introduction}, our inference-time method generates more
uniform object response across the entire object.
We conduct extensive experiments on the PASCAL VOC 2012 dataset \cite{everingham2010pascal}, and both qualitative and quantitative results demonstrate the effectiveness of our approach.

In addition, we explore a new method to leverage saliency maps in WSSS. We argue that, since salient object detection models are always trained by the class-agnostic objects with center bias,
directly using saliency as background cues to generate pseudo labels can deteriorate the quality of the
segmentation 
pseudo labels.
To address this,
we propose \emph{activation aware mask refinement} to further refine the semantic segmentation,
which
uses saliency maps as a subsidiary supervision during semantic segmentation training along with pseudo labels. 
We can refine the foreground object boundaries and meanwhile inhibiting suppression on the activated objects in the background. 

Our main contributions are summarized as follows:
1) We identify a core issue causing the unequal distribution of CAMs and explore a new option to refine initial response maps during inference.
2) We propose a \emph{Class-conditional Inference-time Module} to obtain better object activation without any network modification or re-training.
3) We propose the \emph{Activation Aware Mask Refinement Loss}, a new approach to incorporate saliency information in WSSS that can refine object boundaries, but also prevents suppression of background objects due to the saliency centre-bias. 
4) Our inference-time method can also be treated as an add-on solution to the existing image-level based methods.


\section{Related Work}
\noindent\textbf{Weakly Supervised Semantic Segmentation:}
A large number of WSSS methods have been proposed to achieve a trade-off between labeling efficiency and model accuracy, where the \enquote{weak} annotations can be
image-level labels \cite{huang2018weakly,wei2018revisiting,ahn2018learning,fan2020cian,papandreou2015weakly, wang2020self, chang2020weakly, zhang2020splitting, chen2020weakly, sun2020mining, zhang2020reliability, guo2019mixup, yun2019cutmix, fan2020learning, lee2021anti, yao2021non, wu2021embedded, lee2021railroad}, scribbles \cite{vernaza2017learning,lin2016scribblesup,tang2018regularized}, points \cite{bearman2016s}, or bounding boxes \cite{dai2015boxsup,papandreou2015weakly,khoreva2017simple,li2018weakly,song2019box, sun20203d}.
We mainly focus on image-level label based weakly supervised models.

As the start point, the quality of the initial CAM is important for the semantic segmentation network. Two different methods have been widely studied to obtain a better initial response map, including network refinement based models and the data augmentation and erasing based techniques.

\noindent\textbf{Network Refinement:
}
\cite{wei2018revisiting} adopts dilated convolution with various dilation rates to enlarge the receptive field and transfer 
discriminative information to 
non-discriminative object regions.
\cite{chang2020weakly} performs clustering on image features to generate pseudo sub-category labels within parent classes, which were then used
to train the classification model and generate
CAMs that
covered larger regions of the object.
\cite{zhang2020splitting} introduces discrepancy loss and intersection loss to first mine regions of different patterns, and then merge the common regions of different response maps.


\noindent\textbf{Data Augmentation and Erasing:}
These methods augment or erase input images to force
the network to generate larger response map object coverage.
\cite{wei2017object}
erases highly activated areas from the image and then retrains the classification network to discover new object regions for classification. 
\cite{singh2017hide} divides the image into a grid of patches, then randomly hides patches to force the network to focus on other relevant object parts. 
\cite{hou2018self} leverages two self-erasing strategies to encourage the network to use reliable object and background cues to prohibit attention from spreading to unexpected background regions.
\cite{chang2020mixup, guo2019mixup} utilize mixup data augmentation to calibrate the uncertainty in prediction, they randomly sample an image pair to mix them together and feed into the network,
which forced the model to pay attention to other image regions.
%


In general, all above methods require \enquote{\textbf{re-training}} the classification model
to obtain a refined initial response map. We introduce a new method for initial response map acquisition without re-training.
One recent method that does not re-train is
\cite{fan2020employing}, which directly generates multiple CAMs for each image from the classifier using different input scales, backbones and post-processing. The segmentation model’s robustness is then leveraged to learn from the noisy CAMs, using a learnable per-pixel weighted sum of multiple CAMs.
In this paper, we explore a new alternative option
to refine the initial response maps during the CAM inference stage of a single network
without re-training.

\noindent\textbf{Saliency Assisted WSSS:}
Saliency maps are often adopted in WSSS
\cite{huang2018weakly, hou2018self, fan2020learning, fan2018cian, wei2018revisiting, jiang2019integral, lee2019ficklenet, sun2020mining, wang2018weakly, wei2017object} to serve as 
background cue to generate pseudo labels.
Recently, \cite{wu2021embedded} proposes a pseudo label generation module, which uses saliency maps as background cues and chooses a predefined threshold to retain the activated objects in background.  
\cite{lee2021railroad} directly utilizes saliency maps as supervision during classifier training to constrain object activation.
\cite{yao2021non} proposes potential object mining and Non-Salient Region Masking to explore objects outside salient regions.
However, no existing WSSS methods directly use saliency maps during semantic segmentation training.

\begin{figure}[t!]
   \begin{center}
   {\includegraphics[width=1.0\linewidth]{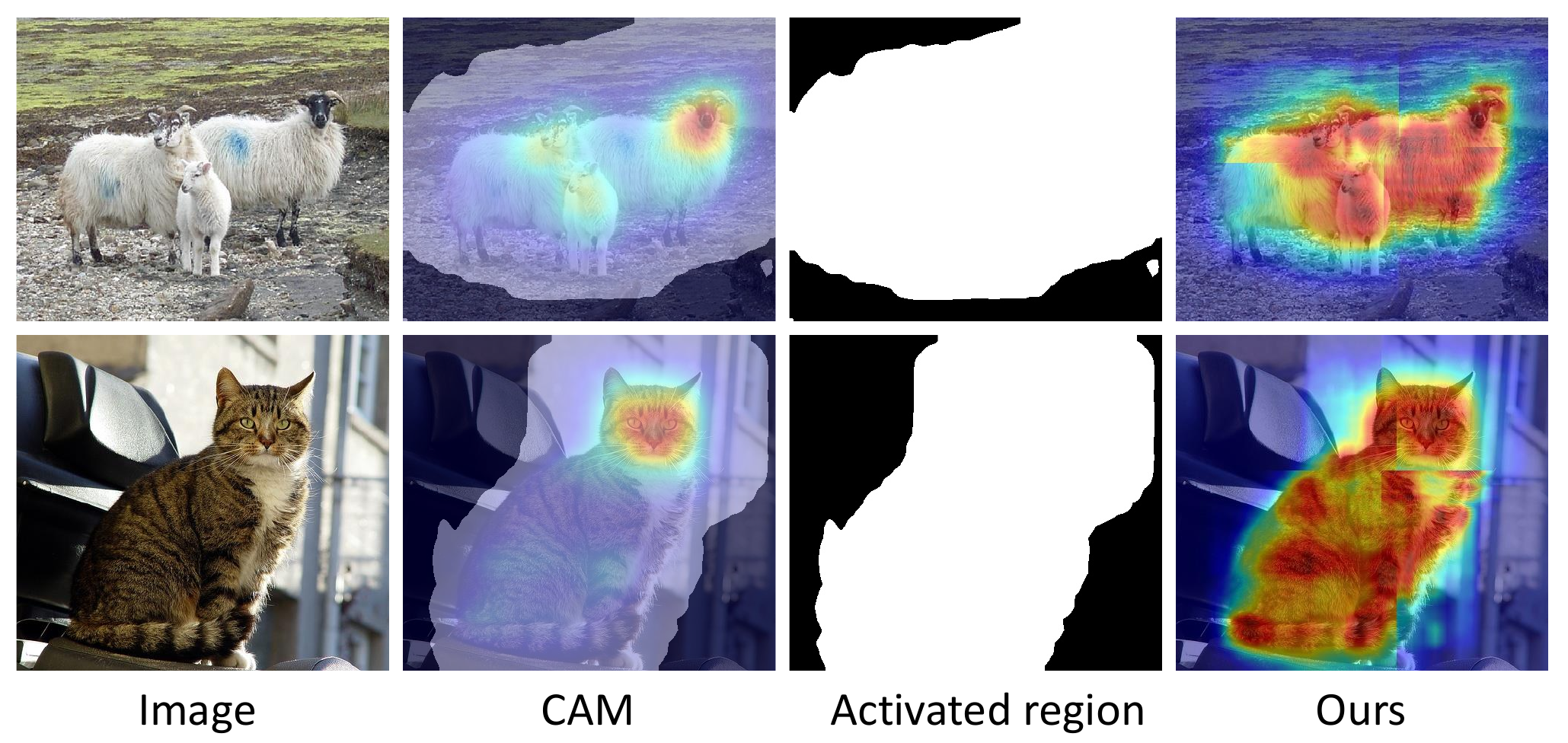}}
   \end{center}
\vspace{-2mm}
\caption{Visualization of activated regions of the baseline CAM. 
Activation is unequally distributed in the object, where the highly activated region (CAM) is the discriminative region.
However, we observe that the
less discriminative regions are still activated covering most object areas (second and third column)}
\vspace{-4mm}
   \label{fig: unequal distribution}
\end{figure}

\section{Our Approach}
We first analyse the baseline CAM method, then propose the \emph{class conditional response map} and a
\emph{class conditional inference-time approach} 
based on it. Finally, we introduce our \emph{Activation Aware Mask Refinement Loss}, a novel approach to leverage pre-trained saliency maps in WSSS.

\subsection{Class Conditional Response Map}
\label{sec: Class Conditional Response Map}
For a class $c$, a CAM \cite{zhou2016learning} is a feature map indicating the discriminative region of the image that influence the classifier to make the decision that this object belong to class $c$.
Given $f_k(x,y)$, the activation of unit $k$ in the last convolutional layer at location $(x,y)$, and $w_c$, the weights from  $f_k(x,y)$ 
via global average pooling, the CAM $M$ is a map of activation at locations $(x,y)$ as described in \cite{zhou2016learning}:
\begin{equation}
M_c(x,y) = \Sigma_k w^c_k f_k(x,y).
\label{ClassicalCAM}
\end{equation}
%

Consider a network $h$ with parameters $\theta$.
For input image $I$, suppose the output for class $c$, $h_{\theta}(c|I) > \tau$, where $\tau$ is the threshold probability. Recent high-performance networks  (e.g., \cite{huang2017densely,He_2016_CVPR}) trained on large datasets (\eg, ImageNet) yield highly accurate classification for well-represented classes.
Hence, given positive classification for $c$, we can treat $M_c$ as an approximation to a class conditional response map.
Hence, we define a \emph{class conditional response map} for image $I$ using network parameters $\theta$ as:
\begin{eqnarray}
    R_c((x,y)|I; \theta)  &= M_c((x,y)|c,I;\theta) \\ &\approx M_c((x,y)|I;\theta), \nonumber
\end{eqnarray}
where $ M_c((x,y)|c,I;\theta)$ is the CAM for $I$ in which  class $c$ appears, and $ M_c((x,y)|I;\theta)$ is the CAM for image $I$, 
given
that class $c$ appears with high probability in the image.

\noindent\textbf{Over-complete Activation}\\
Consider Fig.~\ref{fig: unequal distribution}, column two  shows the baseline CAM activation, where some discriminative areas are well-activated, but other visible object regions have weak activation, (\eg, the heads of the sheep versus the bodies). The third column shows all areas with activation greater than zero, which are large and generally include most object areas. Quantitatively, we obtain all activated masks for the PASCAL VOC 2012 training set and get a recall of 84\% compared with the semantic segmentation ground truth, \ie the majority of object regions are activated by the baseline classifier, not just discriminative regions.
We can see that the baseline classifier learns most object features with sufficient training data.
However, the response is over-complete and uneven,
so extracting segmentation pseudo labels is difficult.

\noindent\textbf{Uneven Distribution}\\
For discriminative training, the loss is indifferent to extensive activation across an object, requiring only a sufficient global average value via pooling.
Existing image-level weakly supervised methods observe that the CAM only shows high activation on an object's most discriminative regions \cite{wei2017object, singh2017hide, wei2018revisiting, hou2018self, chang2020mixup, guo2019mixup, chang2020weakly, wang2020self, zhang2020splitting},
but they disregard the less discriminative object regions where the activation is suppressed.
By definition, the class conditional response map has a global average per-pixel response greater than a threshold. Then, we can infer that a unit $f_k(x,y)$ at location $(x,y)$ with high activation in $R_c((x,y)|I;\theta)$ has an appearance pattern within its receptive field that is strongly associated with the presence of $c$.

\noindent\textbf{Suppression of Broader Object Activation}\\
We argue that for a particular input image, the presence of a highly discriminative region may suppress other less discriminative regions in $R_c$. 
Network processing is well-understood, but let's
consider network mechanisms that feed into $R_c$.
The main mechanism for suppression in earlier layers in modern networks (\eg, ResNet) is via batch normalization and negative weights (ReLU output is non-negative). Each $f_k(x,y)$ from 
Eq.~\ref{ClassicalCAM} projects back into the penultimate layer by standard convolution:
\begin{equation}
f_k(x,y) = \sigma (\Sigma_{k' \in \mathcal{N}_{(x,y)}} [w^c_{k'} f_{k'}(x,y)]),
\label{recursiveinputs}
\end{equation}
where we use $\sigma$ to represent a combined ReLU activation with Batch Normalization, and $k'$ indexes over the convlution  input units to $f_k(x,y)$.
That is, the class conditional response at location $(x,y)$ is a non-linear function of weighted input units $f_{k'}(x,y)$ from the prior layer.
Hence, by a combination of negative $w_{k'}$ and $\sigma$, a strong activation from some $f_{k'}$ can lead to suppression of the corresponding $R_c((x,y)|I'; \theta)$ in the class 
conditional map. Note that $f_{k'}(x,y)$ can also be negative by the cascade of Eq.~\ref{recursiveinputs}. 

For networks with deep residual structures it would be difficult to trace back through the cascading and residual activation to find the exact pixels that have led to this suppression of CAM locations that may otherwise have positive features. 
Instead, we propose a class conditional inference-time approach to solve this issue. We aim to generate a more uniform distribution of activation across the visible object.

\subsection{Class Conditional Response Map Inference}
We assume an initially well-trained classification network $h_\theta$, which we do not re-train. Instead we propose \emph{class conditional inference}, whereby we augment $I$ at inference time to remove regions that may suppress the response of other object parts  to mine the class-conditional object activation. 
Concretely, we perform this by computing the class conditional response map $R_c(x,y|I';\theta)$, where $I'$ is an augmented image. 
By removing pixels in $I$ associated with high activation, we aim to explore other regions of being suppressed. A high response on $R_c(x,y|I';\theta)$ for some location where the activation was not high for  
$R_c(x,y|I;\theta)$ is likely to indicate an appearance pattern that is strongly associated with the object, but was suppressed by the regions that visible in $I$, but not $I'$. In this section we describe our implementations of this by class conditional Split $\&$ Unite Augmentation, and Iterative Inference.

\noindent\textbf{Split $\&$ Unite: Image Augmentation}\\
\label{sec:split}
To address the unequal distribution of object activation, we propose a class conditional split $\&$ unite inference strategy to investigate different parts of the object, and align their activation.
A naive approach would be to randomly divide the image by a grid and perform inference on each grid cell. 
But it is likely that some cells would not
contain the object, and have a higher chance of false positive responses on small regions. 

\begin{figure}[!htb]
   \begin{center}
   {\includegraphics[width=1.0\linewidth]{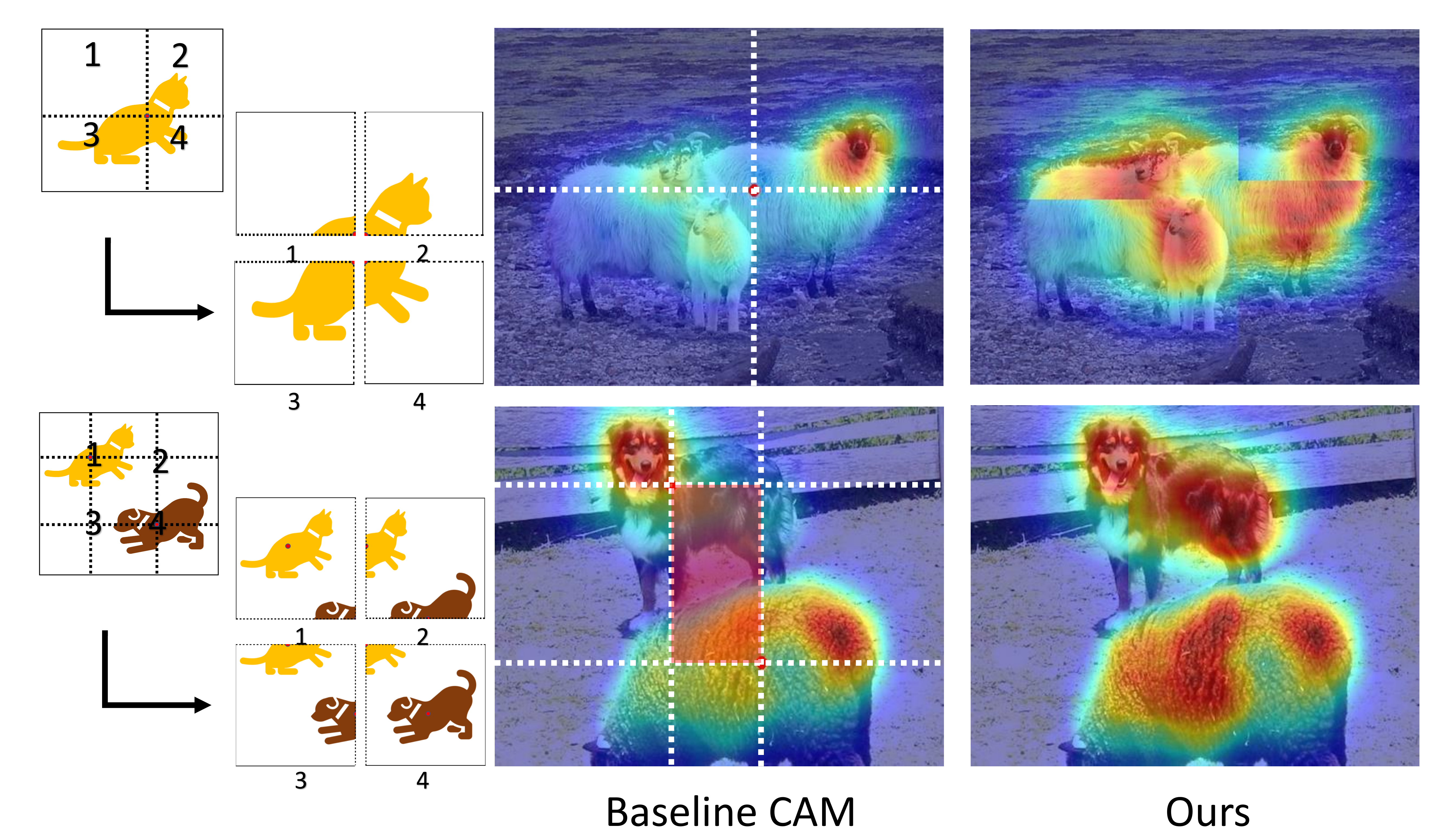}} 
   \end{center}
 \vspace{-3mm}
\caption{Our split $\&$ unite augmentation methods. 
During inference, we split the image into
patches with each patch containing a part of object,
and then we reprocess
each patch to find its
class conditional response map.
}
\vspace{-2mm}
   \label{fig:split}
\end{figure}

Instead, we first apply conventional inference with the baseline classifier to obtain
$R_c((x,y)|I;\theta)$.
As we assume a well-trained model $\theta$, the initial high CAM activation will generally fall on the discriminative region of the target object. 
Then we calculate the centre of mass of the original CAM to obtain a center point about which we split the image into four patches, as shown in Fig.~\ref{fig:split}. We find the centre of mass generally falls on the class, in which case some part of the object appears 
in each of the four sub-images. 

With the classifier fixed, we then compute $R_c((x,y)|I_i;\theta)$ where $i$ indexes each split,
Although the unequal distribution of the object activation in the baseline CAM leads to only highly discriminative regions being highlighted,
by separate inference, object discrimination is computed individually in each patch with any suppressing elements in the other patches removed so we can locate more discriminative areas on different object parts.
For example, see the first row of Fig.~\ref{fig:split} showing the split for images with one object class.
Without the highly activated sheep head in the patch \#2, other parts of the sheep are highlighted in patch \#1, \#3 and \#4.
For images with two object classes, we refine our split strategy,
as shown in second row of Fig.~\ref{fig:split}, we calculate centres of
mass for both class's CAM, then we use these two center points to obtain a rectangle inside the image (displayed as the red central area). 
We use the four corners of this rectangle as a split point to crop the original image into four patches. 
The central rectangle (red area) is retained in all four patches, as shown in the \enquote{Splits} column. Each split generally contains different parts of both object classes.
Then we run inference on each patch in turn, and merge the four response maps, we take the max activation for the overlapping central rectangle area. 
For images with three or more classes, we split the images by the CAM mass center for each class separately.




Furthermore, our split $\&$ unite method could also be used as a common image augmentation method to assist re-training the classification model.
Specifically, we feed each patch individually into the classification model to train the network with the same image-level ground truth as the original image.
\ie we use different object parts to train the classifier instead of the entire object, thus the classifier will naturally pay attention to more object parts during inference.
We show further experimental results of in Sec.~\ref{sec ablation re-train}.




\begin{figure}[t!]
  \begin{center}
  {\includegraphics[width=1.0\linewidth]{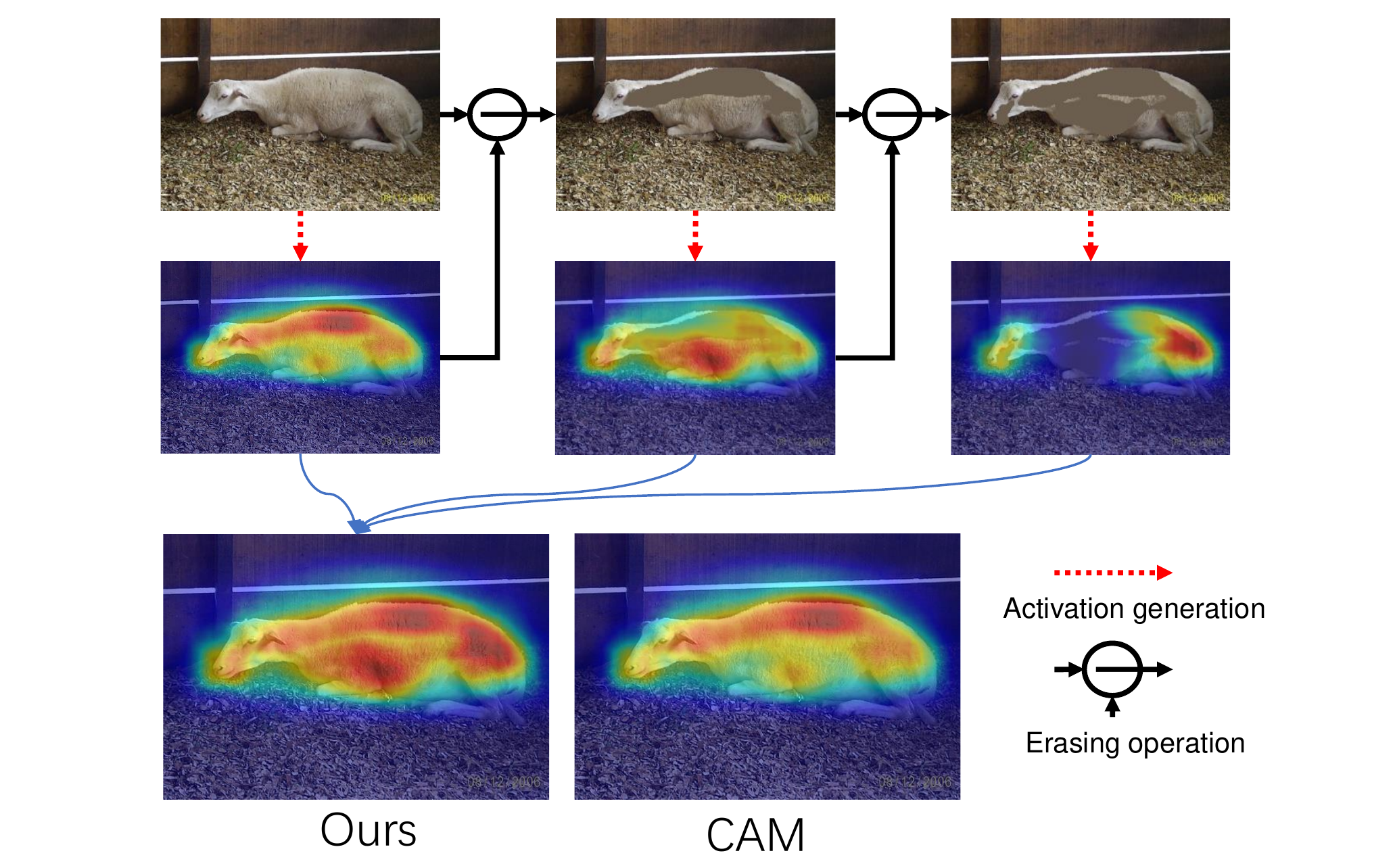}}
  \end{center}
    \vspace{-3mm}
\caption{
We iteratively remove the highly activated regions from the image, and
feed the new image back to the fixed classifier.
}
\vspace{-2mm}
  \label{fig: iterative}
\end{figure}

\noindent\textbf{Refined Iterative Inference Module}\\
In the Split \& Unite module, the classifier still focuses on discriminative regions in each patch,
so we extend our inference method by introducing the iterative erasing mechanism.
Iterative training by erasing the high activation is an adopted technique in the WSSS.
For example, \cite{wei2017object} proposes an iterative training scheme, in each iteration the high activation regions are erased then the image is fed back to train the classifier again.
But it has the risk that all objects are erased while network is still updating with false positive, also the re-training time is greatly increased.
In this section, we propose a refined inference-only iterative module to further improve the object activation maps. Our module is supported by our analysis in Sec. \ref{sec: Class Conditional Response Map}, which
requires no re-training, avoid the risks of existing iterative erasing techniques and can achieve better results.



Refer to Fig.~\ref{fig: iterative}, we first feed the original image into the classifier and produce response map,
revealing 
the highly discriminative areas.
Then these highly discriminative areas are erased with mean colour value of the original image. 
The augmented image is then fed back to the classifier for next
inference iteration. 
No training is performed
as the pixels corresponding to the discriminative region are
removed. With these object features absent, without suppression,
weaker activation will be naturally driven to shift to high activation.
We iterate this inference process and then add the newly generated activation map of every iteration together to obtain the final response map.
As shown in Fig.~\ref{fig: iterative}, new object areas are activated progressively in each iteration without updating the network. 
\begin{figure}[t!]
   \begin{center}
   {\includegraphics[width=1.0\linewidth]{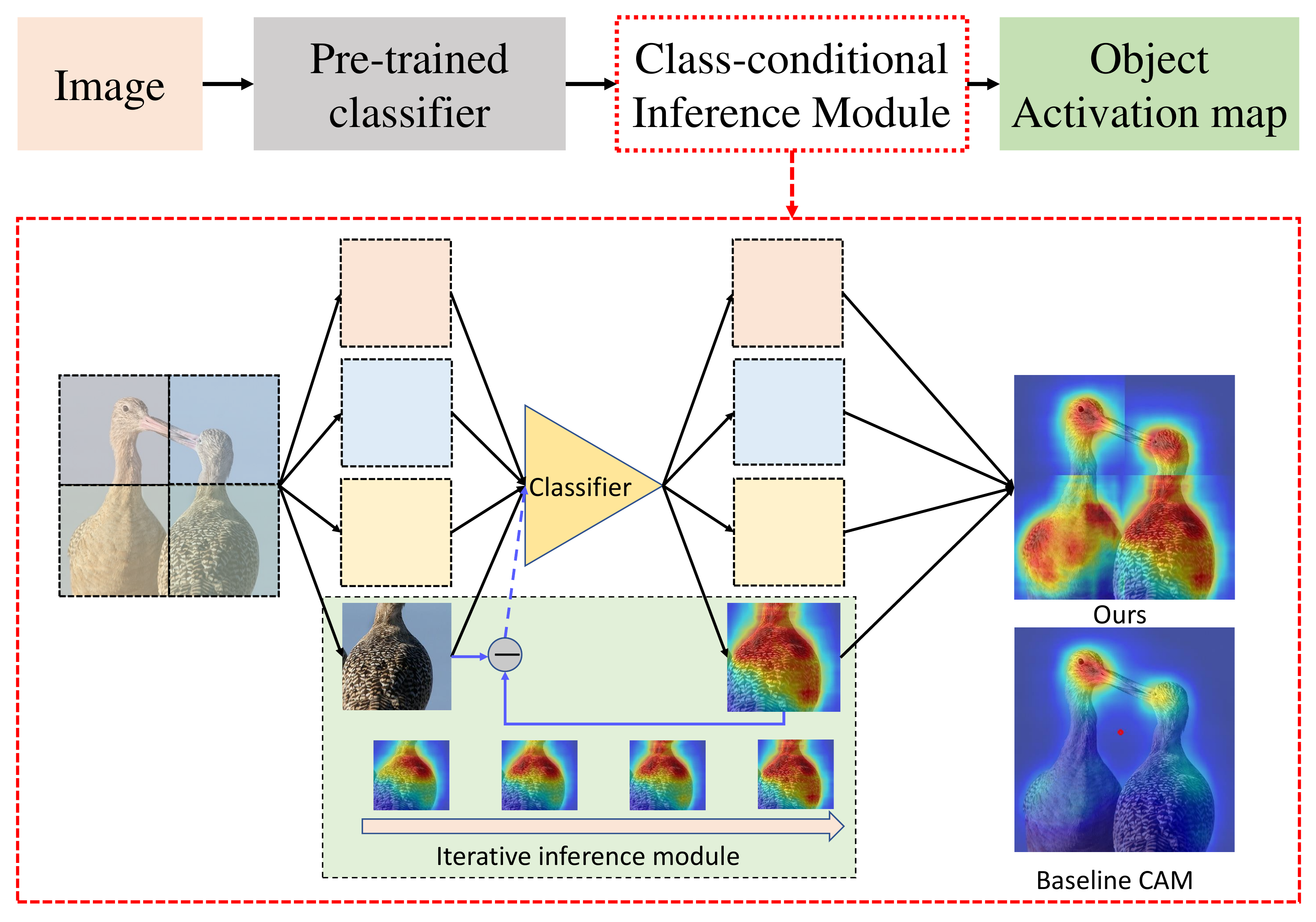}}
   \end{center}
\caption{The framework of our proposed inference method.
As shown, we split the image into 4 splits and do iterative inference(green block) on each split in parallel, then we combine the 4 splits together into our final object response map.
}
\vspace{-2mm}
   \label{fig: framework}
\end{figure}

\noindent\textbf{Our Inference Module}\\
Finally, refer to Fig.\ref{fig: framework}, we integrate our split \& unite and iterative inference together into a unified inference-only framework.
We first split the image into 4 splits and feed them into the classifier in parallel to encourage more object activation on each split.
Then we perform iterative inference on each of them respectively
as shown in the green block.
Finally, we combine each split together into our final object response map. 
The split \& unite module and iterative inference module mutually benefit each other to balance
activation across different object parts and densely cover larger object areas. 
Our class-conditional module can be seemed as an add-on module.
It can be seamlessly utilized in the inference stage of any pre-trained classification networks.
As the proposed inference module requires no re-training, it saves all re-training time and is much easier to implement. 
In our experiments section, we compare with existing methods, and the results show that we achieve comparable performance with the
state-of-the-art methods while most of them
rely on re-training the classification model.

\subsection{Activation Aware Mask Refinement Loss}
Saliency information is used by many approaches in image-level WSSS to refine object boundaries and obtain a background mask
\cite{sun2020mining, jiang2019integral, lee2019ficklenet, fan2020cian, fan2018associating, hou2018self, huang2018weakly, wei2018revisiting, wang2018weakly}.
Commonly, a pre-trained off-the-shelf salient object detection model is adopted to generate class-agnostic saliency masks on the segmentation dataset. 
Then the saliency masks are used during pseudo label generation, normally they
are multiplied by a manually chosen parameter and concatenated onto the activation maps as background. High activation pixels are then used to obtain pixel-wise 
pseudo labels.

\begin{figure}[!t]
   \begin{center}
   {\includegraphics[width=0.90\linewidth]{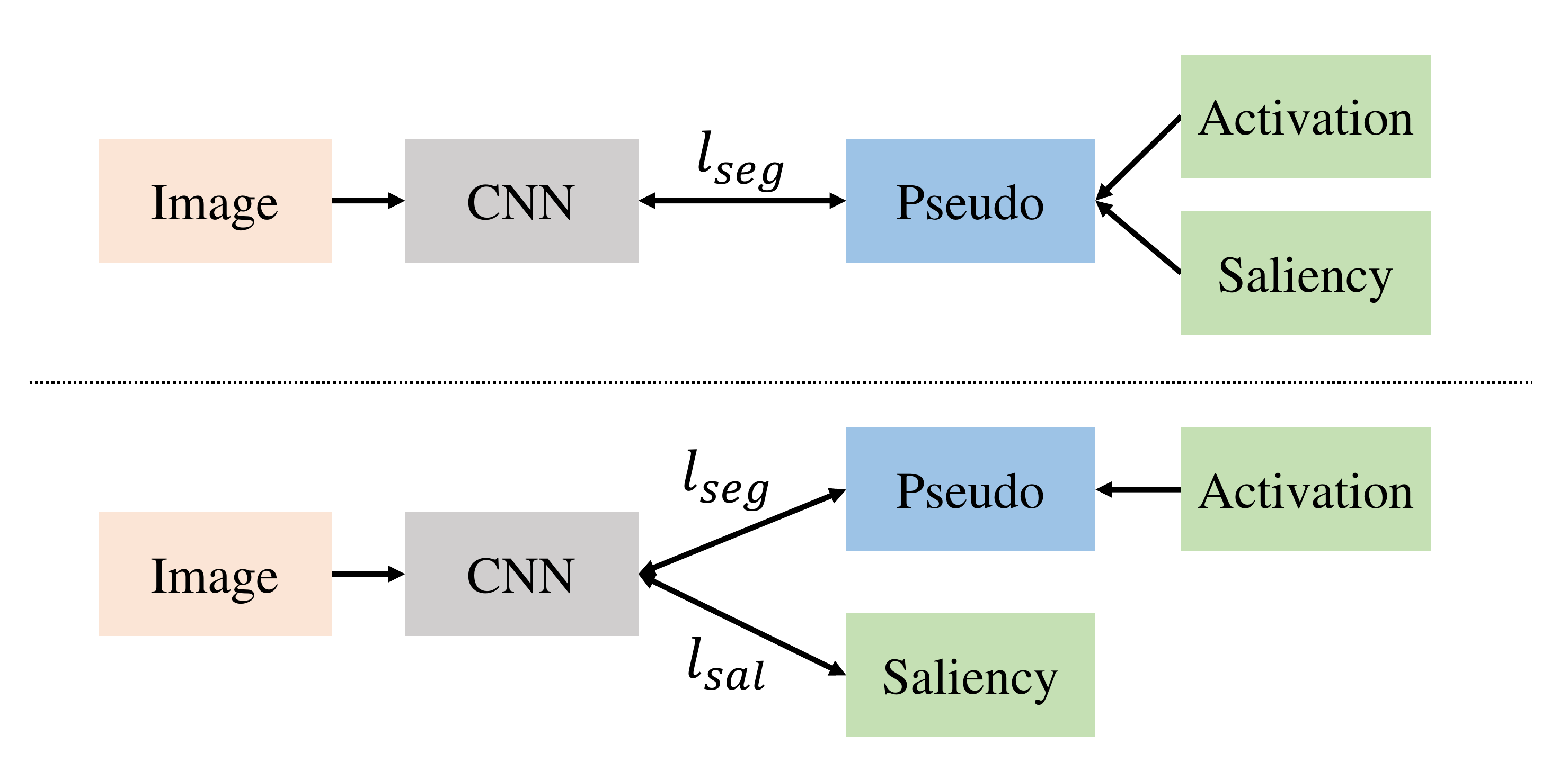}}
   \end{center}
\caption{Top: the convention saliency guided WSSS training.
Bottom: our activation-aware mask refinement training, where object activation and saliency information are visible to the network.
}
\vspace{-2mm}
   \label{fig: saliency model}
\end{figure}
\begin{figure}[!htb]
   \begin{center}
   {\includegraphics[width=0.90\linewidth]{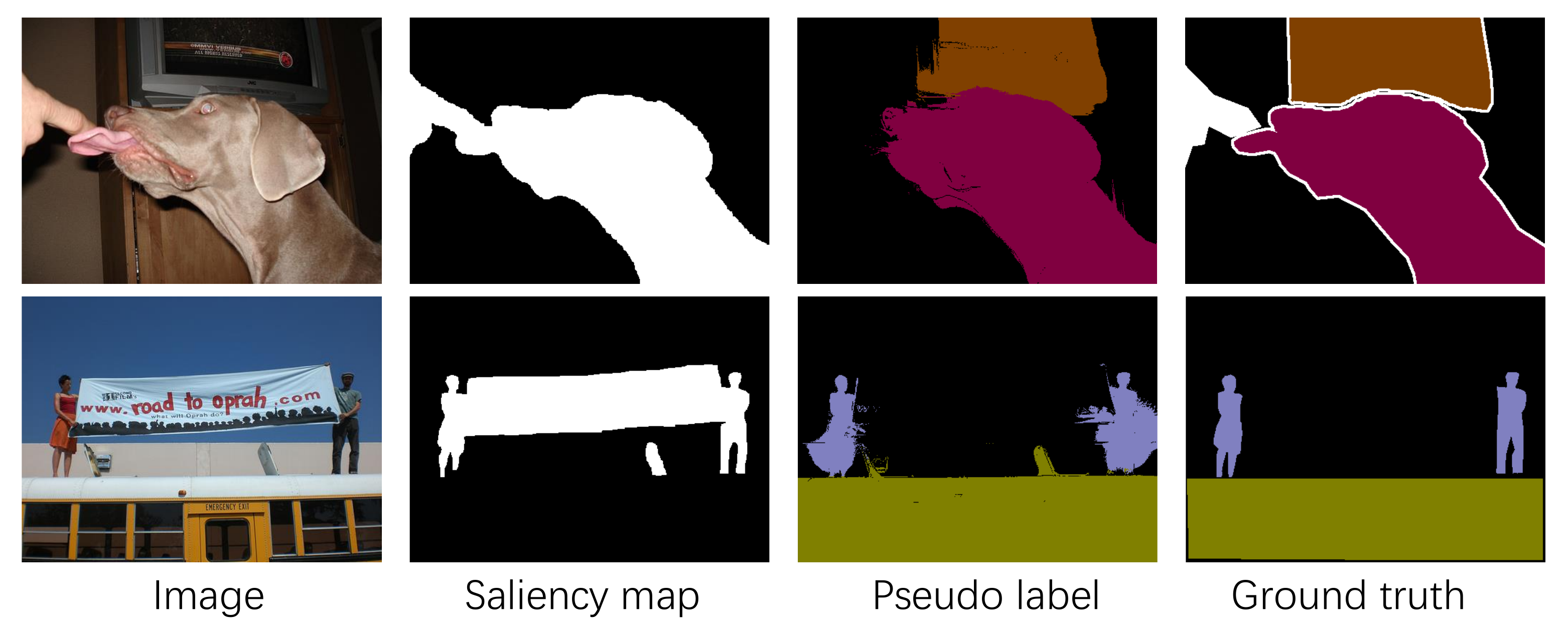}}
   \end{center}
\caption{Conflicts between saliency maps and our activation-generated pseudo labels, where the \enquote{Pseudo label} is the
activation-generated pseudo label.
As shown in the first row, our activation correctly detects the TV monitor but the saliency map 
ignores it.
In the second row, the saliency map falsely highlights the banner but ignores the bus.}
   \label{saliency_fig}
\end{figure}

However, we observe that since saliency detection models are usually trained by class-agnostic objects with center bias,
the saliency maps may falsely detect non-object salient areas in the foreground, and tend to
ignore non-salient objects in the background (see Fig.~\ref{saliency_fig}),
Thus, it may introduce errors into pseudo-labels and harm segmentation training. 
Some desired object regions are ignored and some non-object regions are falsely detected in the saliency maps.
However, saliency maps are still required by WSSS to support finding accurate foreground object boundaries.
To address this issue, we propose a new method to leverage  saliency information in WSSS, including a new saliency loss.
The use of CAMs to obtain pseudo-labels without saliency maps is unchanged.
Then we use the
pseudo labels together with the provided saliency maps to 
train our semantic segmentation model.
As shown in Fig.~\ref{fig: saliency model},
we keep both labels visible to the network, 
so the saliency maps can refine foreground object boundaries, 
but we inhibit suppression of 
activated objects in the background.
More formally, our segmentation loss is defined as:

\begin{equation}
\label{eq: seg_loss}
    L_{seg} = L_{seg} + \alpha (e^{\tau}-1) L_{sal} 
\end{equation}
\vspace{-3mm}


The first term $L_{seg}$ is the cross entropy loss between segmentation prediction and our activation-generated pseudo labels.
The second term $L_{sal}$ denotes the binary cross entropy loss between the background channels of the predictions and the saliency maps.
We incorporate a modulating factor $\tau$ (called the \emph{conflict temperature}) with tunable weight $\alpha > 0$.
Intuitively, as two noisy signals (pseudo labels and saliency maps) in the supervision pair have conflict areas as shown in Fig.~\ref{saliency_fig}, they will compete with each other.
We use $\tau$ to control the competition between two supervision terms. $\tau$ is
defined as the mean intersection-over-union (MIoU) between our pseudo label background channel and 
the saliency maps.
If  $\tau$ is
low, it indicates the saliency map is not consistent with the pseudo background, and so the saliency maps will harm  segmentation training,
so the weight of the saliency loss $L_{sal}$ is diminished. 
Contrarily, if MIoU is high, then the saliency loss is enhanced to refine the object boundaries.
In summary, our activation-aware mask refinement loss proposes an adaptive mechanism to refine segmentation training.
Saliency information is fully explored to refine the foreground object boundaries but not harm activated objects in the background.

\section{Experiments}
\subsection{Implementation Details}
In this section, we introduce implementation details of the proposed method and the following procedures to generate semantic segmentation results.
We evaluate our approach on the PASCAL VOC 2012 dataset \cite{everingham2010pascal} with the background and 20 foreground object classes.
The official dataset consists of  1446 
training
images,
 1449 
 validation and 1456 
test. 
We follow common practice, augmenting the training set by adding images from the SBD dataset \cite{hariharan2011semantic}, to form a total
of 10582 images.

\begin{figure*}[!htb]
   \begin{center}
\includegraphics[width=0.70\linewidth]{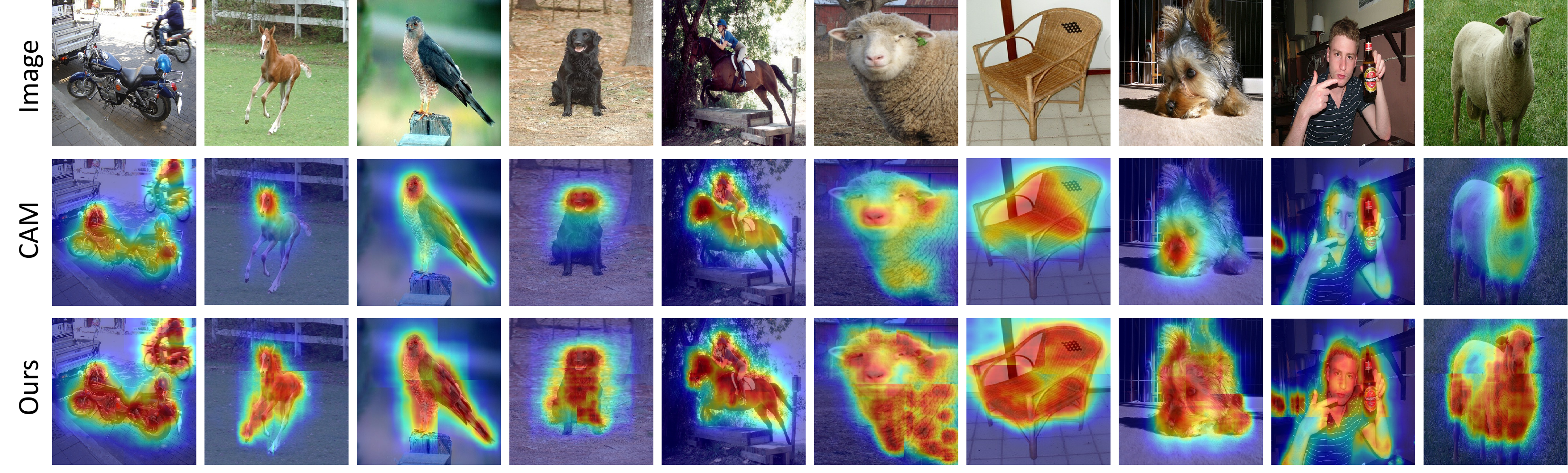} 
   \end{center}
\caption{Sample results of initial response maps and ours. Baseline CAMs tend to only highlight discriminative areas.
Our approach helps balance object activation across different object parts and densely cover larger object areas.
For images with multiple classes, we can get better activation for all classes, and we combine them in the activation for convenience of reading.
}
\vspace{-3mm}
   \label{fig:results}
\end{figure*}

\noindent\textbf{Object Response Map Generation}\\
In our pipeline, we use the weights of the baseline classification model provided by \cite{ahn2018learning}, pre-trained on 
ImageNet \cite{deng2009imagenet}, and fine-tuned on PASCAL VOC 2012, without any re-training.
Similar to \cite{ahn2018learning} and others,
our baseline classifier uses the ResNet-38 backbone with output\_stride = 8, global average pooling, followed by a fully connected layer. 
In the iterative inference module, we empirically choose 0.7 as threshold for high activation, and remove high activation regions in each inference iteration. 
We stop iteration when the new
activation is smaller than 1\% of the image. 


\begin{table}[]
\footnotesize
\centering
\begin{tabular}{ccc|c}
\hline
                               & \multicolumn{2}{c|}{Training Set} & Validation Set \\ \hline
\multicolumn{1}{l|}{Method}    & CAM           & CAM + RW          & CAM            \\ \hline
\multicolumn{1}{l|}{Baseline \cite{ahn2018learning}}  & 48.0          & 58.1              & 46.8           \\ 
\multicolumn{1}{l|}{Mixup-CAM \cite{chang2020mixup}} & 50.1          & 61.9              & x              \\ 
\multicolumn{1}{l|}{SC-CAM \cite{chang2020weakly}}        & 50.9          & 63.4              & 49.6           \\ 
\hline
\multicolumn{1}{l|}{Ours}      & 52.2          & 64.2              & 50.7             \\ \hline
\end{tabular}
\caption{Performance comparison in mIoU(\%) of the initial response maps on the PASCAL VOC training and validation set. Mixup-CAM \cite{chang2020mixup} and SC-CAM \cite{chang2020weakly} are two
state-of-the-art methods, and
x means the paper does not report this number.}
\label{table cam}
\end{table}

\begin{table}
\footnotesize
\centering
\begin{tabular}{ccc|c}
\hline
Baseline & split \& unite & Iterative Inference & mIoU \\ \hline
\checkmark &     &                 &            48.0              \\
\checkmark &  \checkmark     &                &         49.9                \\
\checkmark &      &           \checkmark      &         50.2                \\ 
\checkmark &   \checkmark   &       \checkmark          &             52.2          \\
           \hline
\end{tabular}
\caption{The ablation study for each part of our method to validate effectiveness of the proposed strategies.
}
\vspace{-5mm}
\label{table source}
\end{table}

\noindent\textbf{Semantic Segmentation Generation}\\
After obtaining response maps using our method, following recent work \cite{wang2020self, chang2020weakly, chang2020mixup, sun2020mining, chen2020weakly}, 
we adopt the AffinityNet random walk \cite{ahn2018learning} to refine response maps into pixel-wise semantic segmentation pseudo labels.
Also, we apply fully connected random fields \cite{krahenbuhl2011efficient} to refine the pseudo-label object boundaries.
Finally, we use the generated pseudo labels and saliency maps as supervision to train the popular Deeplab semantic segmentation framework with ASPP \cite{chen2017deeplab} using the ResNet-101 backbone network. 
We use  saliency maps from \cite{jiang2019integral}, $\alpha$ in Eq.~\ref{eq: seg_loss} is empirically set to 0.08.

\subsection{Improvement on Initial Response Map}
In Table~\ref{table cam}, we show initial response map (CAM) performance, using best mean IoU,
\ie the best match between the response map and segmentation ground truth under all different background thresholds.
We also report pseudo-label results 
after applying the random walk refinement (CAM + RW). 
Note
that our results are obtained with the baseline classifier without any re-training, all improvements come from our proposed \emph{class conditional inference}.
As shown in Table~\ref{table cam}, our initial response maps are significantly improved over baseline \cite{ahn2018learning} on both training and validation sets. We also compare
response maps generated by recent state-of-the-art methods \cite{chang2020mixup,chang2020weakly}, and observe a
a clear
margin. 

Our better initial response maps lead to better performance of downstream tasks: generating pixel-wise pseudo-labels and final segmentation results.
After refining the response map into  pseudo labels that are used to train a semantic segmentation model,  
we also achieve significant improvement over the baseline and outperform competing methods, as shown in \enquote{CAM+RW} 
column in Table~\ref{table cam}.
This validates that, by direct inference without fine-tuning, our method can substantially improve object activation and generate better response maps than competing methods with re-trained classification models.
In Fig.~\ref{fig:results}, we show qualitative examples compared with baseline CAMs, showing that ours can activate substantially more object parts and
uniformly cover larger regions of the objects.

\begin{table}[]
\centering
\footnotesize
\begin{tabular}{c|cccc}
\hline
     Class number & 1 class & 2 classes & 3+ classes & Total \\ \hline
Baseline &  46.6 & 50.9 & 41.4 & 48.0  \\ 
SC-CAM\cite{chang2020weakly} &  49.5 & 53.5 & 43.5 & 50.9 \\ 
Ours & 49.8 & 55.3 & 44.7 &52.2       \\ \hline
\end{tabular}
\caption{Model performance with respect to number of categories.
}
\vspace{-2mm}
\label{table per class cam}
\end{table}

In Table~\ref{table source}, we show an ablation study, how each of our modules improves the initial response maps.
The improvement mainly stems from more dense coverage of objects, we test each module independently. 
Our split $\&$ unite augmentation improves the baseline CAM by 1.9\%. 
The iterative inference module has an improvement of 2.2\% compared with baseline.
It validates that both our modules provide manifest improvement over the baseline CAM.  Integrating them together, we achieve a significant improvement over the baseline by 4.2\% on our initial response map.  

Finally, 
we report performance improvements for different numbers of classes appearing in an image in Table~\ref{table per class cam}.
As shown, we achieve consistent improvements over competing methods on images with all class numbers,  demonstrating the effectiveness and generality of our method. 



\begin{table}[]
\footnotesize
\centering
\begin{tabular}{l|c|c}
\hline
         & Training Set & Validation Set \\ \hline
Baseline & 48.0         & 46.8           \\
Ours (Direct inference)  & 52.2         & 50.7           \\
Ours + split\&unite re-training  & 52.7         & 51.8             \\
Ours + SC-CAM\cite{chang2020weakly}  & 53.3         & 51.5             \\
\hline
\end{tabular}
\caption{The effectiveness of our model as an \enquote{add-on} to
re-trained classifier.
}
\vspace{-5mm}
\label{table retrain}
\end{table}

\subsection{Ablation: Integrating with Re-training}
\label{sec ablation re-train}
Our method can be directly used as an add-on inference module with any existing re-trained classifier to obtain a better initial response map.
First, as discussed in Sec.~\ref{sec:split},
our split $\&$ unite augmentation can be used in the training stage as a
data augmentation method. We follow the methods described in Sec.~\ref{sec:split} to augment the training set to
feed different parts of the object into the classification model to train the network, so the network will update its weights to pay attention to more object parts.
Refer to Table~\ref{table retrain}: fine-tuning the classifier with our split $\&$ unite augmentation further improves the initial response map performance.

In addition, we perform our inference-time data augmentation on the re-trained CAM of (SC-CAM \cite{chang2020weakly}) to refine their produced CAM. SC-CAM \cite{chang2020weakly} introduced a sub-category clustering method to force the network to activate in more categories, leading to enlarged activation regions. We then perform our inference-time approach on their
re-trained classifier.
As shown in Table~\ref{table retrain},
we (Ours + SC-CAM) improve their performance significantly by 2.4\%.
This validates that our inference method can be used as an add-on solution to integrate with existing re-trained classifiers to further refine the object activation maps.
.




\begin{figure}[t!]
   \begin{center}
   {\includegraphics[width=0.90\linewidth]{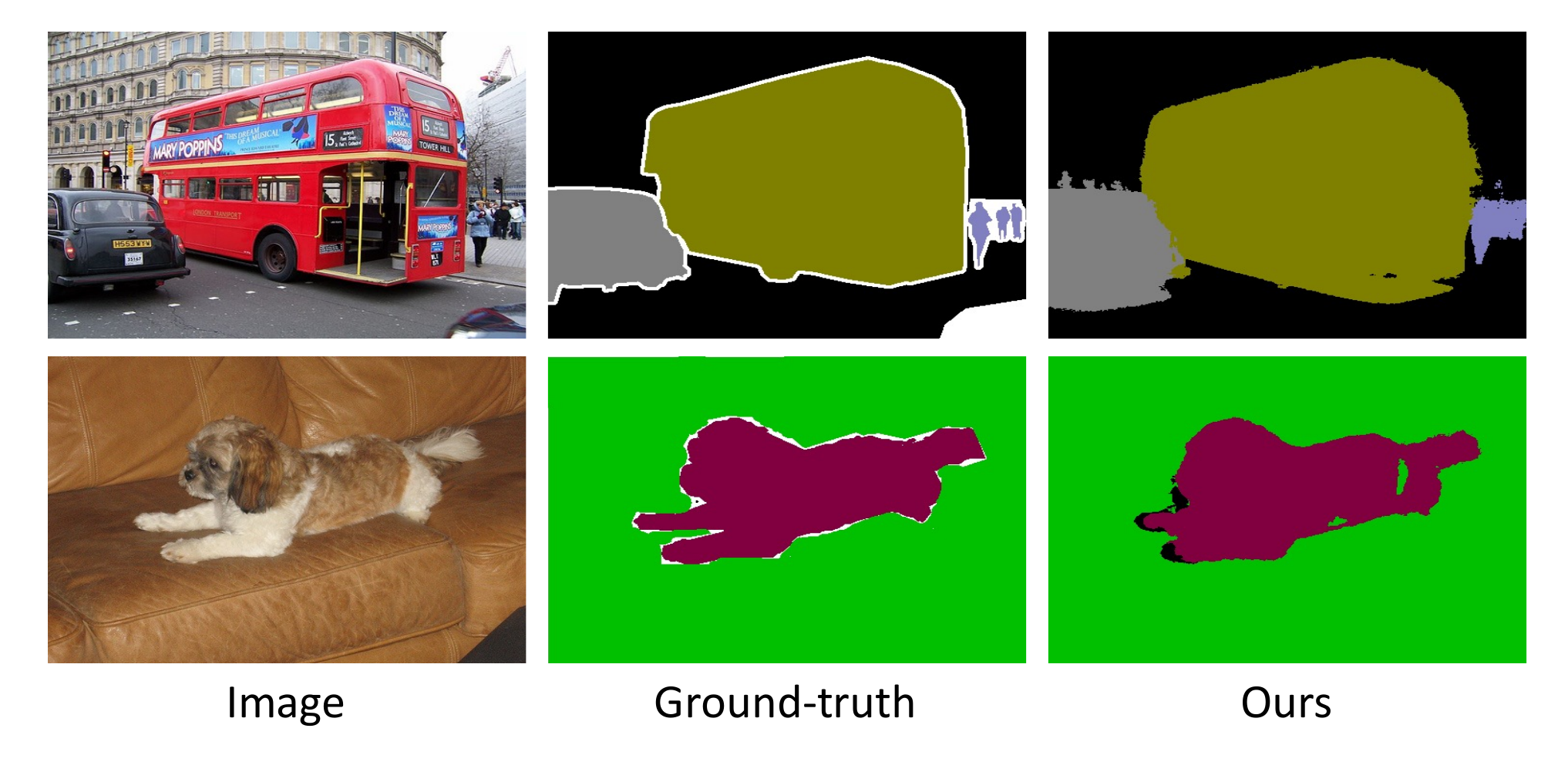}}
   \end{center}
\vspace{-2mm}
\caption{Qualitative results of our semantic segmentation results.}
\vspace{-2mm}
   \label{fig: seg}
\end{figure}

\subsection{Semantic Segmentation Performance}
In Table~\ref{table cam}, we show the refined pseudo labels generated by our initial response maps.
We then use the pseudo labels and our \emph{activation aware mask refinement loss} to train a segmentation network on PASCAL VOC dataset.
As per common practice, we use the densely connected CRF \cite{krahenbuhl2011efficient} to refine the semantic segmentation predictions as post processing. We show final predictions in Fig.~\ref{fig: seg},  clearly showing the effectiveness of our approach. 
In Table~\ref{table sota}, we compare our method with recent work.
We report performance on both the validation and test set of the PASCAL VOC 2012 dataset \cite{everingham2010pascal}. 
As shown, 
our method outperforms all others on the test set, and is comparable with state-of-the-art on the validation set.
Note that most other methods require additional re-training steps
to obtain better object activation,
our method is easier and faster to implement. Further, we remove the saliency guidance in Fig.~\ref{fig: saliency model} from our framework, leading to the cheapest model (no re-training, no saliency), and achieve mIoU(\%) on PASCAL VOC validation set and testing set as 66.2 and 66.3 respectively, which is comparable with re-training based model RRM~\cite{zhang2020reliability} and slightly worse than AdvCAM \cite{lee2021anti}.

\begin{table}[t]
\footnotesize
\begin{tabular}{lllll}
\hline
Method & \multicolumn{1}{p{10mm}}{Re-training}  & \multicolumn{1}{p{10mm}|}{Extra data} & Val & Test \\ \hline
  MDC\cite{wei2018revisiting}        \scriptsize{VGG16}       &     \multicolumn{1}{c}{\checkmark}     &    \multicolumn{1}{c|}{S} &  60.4   &  60.8       \\
    DSRG\cite{huang2018weakly}       \scriptsize{ResNet101}       &     \multicolumn{1}{c}{\checkmark}     &    \multicolumn{1}{c|}{S} &  61.4   &  63.2       \\
 Affinity\cite{ahn2018learning}            \scriptsize{ResNet38}   &     \multicolumn{1}{c}{}     &    \multicolumn{1}{c|}{} &  61.7   &  63.7     \\
  SeeNet \cite{hou2018self}          \scriptsize{ResNet101}    &     \multicolumn{1}{c}{\checkmark}     &    \multicolumn{1}{c|}{S} &  63.1   &62.8      \\
   IRNet \cite{ahn2019weakly}         \scriptsize{ResNet50}    &     \multicolumn{1}{c}{\checkmark}     &    \multicolumn{1}{c|}{} &  63.5   &64.8      \\
    BDSSW \cite{fan2018associating}      \scriptsize{ResNet101}       &     \multicolumn{1}{c}{}     &    \multicolumn{1}{c|}{S}&  63.6   &    64.5  \\
   CIAN \cite{fan2020cian}         \scriptsize{ResNet101}     &     \multicolumn{1}{c}{\checkmark}     &    \multicolumn{1}{c|}{S} & 64.1    &  64.7    \\
   SEAM\cite{wang2020self}             \scriptsize{ResNet38} &     \multicolumn{1}{c}{\checkmark}     &    \multicolumn{1}{c|}{} &  64.5   &     65.7 \\
FickleNet\cite{lee2019ficklenet}           \scriptsize{ResNet101}     &     \multicolumn{1}{c}{\checkmark}     &    \multicolumn{1}{c|}{S} &  64.9   &  65.3    \\
    OAA \cite{jiang2019integral}        \scriptsize{ResNet101}    &     \multicolumn{1}{c}{\checkmark}     &    \multicolumn{1}{c|}{S} &  65.2   &     66.4 \\
       Mixup-CAM \cite{chang2020mixup}          \scriptsize{ResNet101}     &     \multicolumn{1}{c}{\checkmark}     &    \multicolumn{1}{c|}{} &  65.6   &  x      \\   
BES \cite{chen2020weakly}          \scriptsize{ResNet101}      &     \multicolumn{1}{c}{\checkmark}     &    \multicolumn{1}{c|}{} &  65.7   &  66.6    \\

  SC-CAM \cite{chang2020weakly}          \scriptsize{ResNet101}    &     \multicolumn{1}{c}{\checkmark}     &    \multicolumn{1}{c|}{} &  66.1   &  65.9      \\   
  CONTA\cite{zhang2020causal} \scriptsize{ResNet101}    &     \multicolumn{1}{c}{\checkmark}     &    \multicolumn{1}{c|}{} &  66.1   &  66.7\\
  MCS \cite{sun2020mining}     \scriptsize{ResNet101}&     \multicolumn{1}{c}{\checkmark}     &    \multicolumn{1}{c|}{S} &  66.2   &  66.9      \\   
    RRM\cite{zhang2020reliability}(two step)       \scriptsize{ResNet101}     &     \multicolumn{1}{c}{\checkmark}     &    \multicolumn{1}{c|}{} & 66.3    &  66.5   \\
       EME\cite{fan2020employing}       \scriptsize{ResNet101}     &     \multicolumn{1}{c}{}     &    \multicolumn{1}{c|}{S} & 67.2    &  66.7   \\
         ICD\cite{fan2020learning}       \scriptsize{ResNet101}     &     \multicolumn{1}{c}{\checkmark}     &    \multicolumn{1}{c|}{S} & 67.8    &  68.0\\
          AdvCAM \cite{lee2021anti} \scriptsize{ResNet101}     &     \multicolumn{1}{c}{\checkmark}     &    \multicolumn{1}{c|}{} & 68.1    &  68.0\\
        NSRM\cite{yao2021non}       \scriptsize{ResNet101}     &     \multicolumn{1}{c}{\checkmark}     &    \multicolumn{1}{c|}{S} & 68.3    &  68.5\\
        EDAM\cite{wu2021embedded}  \scriptsize{ResNet101}     &     \multicolumn{1}{c}{\checkmark}     &    \multicolumn{1}{c|}{S} & 70.9    &  70.6\\
        EPS\cite{lee2021railroad}  \scriptsize{ResNet101}     &     \multicolumn{1}{c}{\checkmark}     &    \multicolumn{1}{c|}{S} & 70.9    &  70.8\\
  \hline
        Ours \scriptsize{ResNet101} &     \multicolumn{1}{c}{}     &    \multicolumn{1}{c|}{S} &  70.8   & 71.8     \\ 
  \hline
\end{tabular}
\caption{Semantic segmentation performance comparison on the PASCAL VOC 2012 val and test sets. We report the methods that re-train the classification model for better response maps.
Also, the methods that utilize extra saliency masks to generate pseudo labels are marked with \enquote{S}.
}
\vspace{-2mm}
\label{table sota}
\end{table}

\subsection{Discussion: Computation-time}
We argue that current WSSS multi-training schemes are complicated and inelegant, so we propose to improve WSSS performance without further increasing computation time.
First, our saliency loss is a subsidiary supervision calculating the binary cross entropy loss during segmentation training, the increased time complexity is negligible, and most SOTA approaches already incorporate saliency to generate pseudo labels. 
Second, although re-training time is saved,
our \emph{class conditional inference module} requires once-off additional inference-time computation to generate pseudo-labels,
we give a quantitative analysis here. 
To obtain activation maps on the PASCAL VOC train set (1464 images) our method takes 40 minutes compared to 13 (factor of 3, \eg, 
an extra 3.5 hours for the 10582 image Augmented dataset).
On the other hand, re-training the baseline classifier as performed by most others 
requires 6 hours on the same GPU settings (note that competing methods have additional augmented images, and/or network enhancements meaning their re-training takes at least this long). 
Thus, our inference-time method still greatly saves overall time to obtain high-quality activation maps.
As future work, we will refine our method to further reduce the inference time.

\vspace{-2mm}
\section{Conclusion}
In this paper, we propose a novel inference method that helps generate better object response maps without re-training the baseline classifier, and a new method to utilize saliency information in WSSS.
Specifically,
we propose two inference-time modules to generate dense
object response maps.
Firstly, we develop an augmentation method and let the classifier
inference on different image parts individually so as to shift the activation to more object areas.
Secondly, we propose an iterative inference that encourages the classifier to progressively mine more object parts by hiding high activation areas during inference.
Whereas
most current state-of-the-art methods 
require multiple training steps, our method directly generates response maps using the baseline classifier.
We show that our algorithm produces a better initial response map with less computation.
In addition, our activation aware mask refinement loss provides a new way to incorporate saliency information in WSSS
which further improves final semantic segmentation performance.


{\small
\bibliographystyle{ieee_fullname}
\bibliography{egbib}
}

\end{document}